\documentclass[11pt,a4paper]{article}
\usepackage[hyperref]{acl2020}
\usepackage{times}
\usepackage{latexsym}
\usepackage{graphics}
\usepackage{graphicx}
\usepackage{multirow}

\usepackage{microtype}

\aclfinalcopy 

\title{Knowledge-based Extraction of Cause-Effect Relations from Biomedical Text}

\author{Sachin Pawar \\
TCS Research \\
  \texttt{sachin7.p@tcs.com} \\\And
  Ravina More\\
  TCS Research \\
  \texttt{ravina.m@tcs.com} \\\And
  Girish K. Palshikar\\
  TCS Research\\
  \texttt{gk.palshikar@tcs.com} \\\AND
  Pushpak Bhattacharyya \\
  IIT Bombay\\
  \texttt{pb@cse.iitb.ac.in} \\\And
  Vasudeva Varma \\
  IIIT Hyderabad\\
  \texttt{vv@iiit.ac.in}
  }

\date{}

\begin{document}
\maketitle
\begin{abstract}
We propose a knowledge-based approach for extraction of {\em Cause-Effect} (CE) relations from biomedical text. 
Our approach is a combination of an unsupervised machine learning technique to discover causal triggers and a set of high-precision linguistic rules to identify cause/effect arguments of these causal triggers. 
We evaluate our approach using a corpus of 58,761 Leukaemia-related PubMed abstracts consisting of 568,528 sentences. We could extract 152,655 CE triplets from this corpus where each triplet consists of a cause phrase, an effect phrase and a causal trigger. As compared to the existing knowledge base -- SemMedDB~\cite{kilicoglu2012semmeddb}, the number of extractions are almost twice. Moreover, the proposed approach outperformed the existing technique SemRep~\cite{rindflesch2003interaction} on a 
dataset of 500 sentences.
\end{abstract}

\section{Introduction}
The immense text present in the biomedical domain is growing day by day in the form of research papers, case reports, patient health records, health related Question-Answering (QA) forums and even social media. The effective extraction of knowledge from this text is key to find solutions to pressing problems in the medical domain such as cancer~\cite{zhu2013biomedical}. Given the scale of this text, it is important to extract this knowledge automatically and store in some machine-readable knowledge representation (e.g., tables or graphs) so that the knowledge can be indexed, queried, analyzed, or inferred further to generate new knowledge. The state of the art text mining systems are error prone due to the challenges and complexity of Natural Language Processing (NLP). Efficient text mining algorithms will help to improve the knowledge extraction from biomedical text and will be of real value in developing knowledge discovery graphs, QA systems and knowledge summarization.

Cause-Effect relations that denote causal dynamics between entities (e.g., {\small\tt Bortezomib causes proteasome}, {\small\tt Shp-2 is upregulated by p210 bcr/abl oncoprotein}), capture critical knowledge about the domain. Such knowledge can be utilized for correctly answering questions like -- {\small\tt What are the causes for apoptosis of Kasumi-1 cells?}, {\small\tt Which drugs given to leukaemia patients cause anaemia as a side effect?} and {\small\tt Tell me all causes for cytotoxicity in tumor cells.}

While popular biomedical knowledge bases such as SemMedDB 
contain causal predicates such as {\small\sf CAUSES}, {\small\sf INHIBITS}, {\small\sf STIMULATES}, etc., from biomedical papers, they are not able to capture all the causal relations. In this paper, we present an approach to augment the causal predications of SemMedDB through an knowledge-based method by extracting more CE relations. 
Our proposed approach is a combination of an unsupervised machine learning technique to discover causal triggers and a set of high-precision linguistic rules to identify cause/effect arguments of these causal triggers. We also use simple rules to extract additional arguments of the CE relations: {\em negation} and {\em uncertainty}. In our experiments with a Leukaemia-related subset of 58,761 PubMed citations, our approach is able to extract 152,655 Cause-Effect triplets whereas for the same subset of citations, SemMedDB has only 77,135 causal predications. Also, the precision of our CE triplets which are extracted over and above SemMedDB, was evaluated to be around 60\% using a random subset.

\section{Cause-Effect Relation Extraction}
We represent a Cause-Effect relation mention in the form of a triplet which consists of:

\noindent $\bullet$ \textbf{Causal trigger}: A multi-word expression or a verb which invokes a Cause-Effect relation, e.g., {\small\tt because}, {\small\tt due to}, {\small\tt causes}, {\small\tt inhibits}.\\
$\bullet$ \textbf{Cause phrase}: A noun or verb phrase which represents a {\em cause} argument of the Cause-Effect relation invoked by the causal trigger.\\
$\bullet$ \textbf{Effect phrase}: A noun or verb phrase which represents an {\em effect} argument of the Cause-Effect relation invoked by the causal trigger.

Consider the following sentence: {\small\tt MMuLV infection of non-transgenic mice induced primarily mature T cell lymphomas.} For this sentence, following CE triplet is extracted:\\
$\langle$\textbf{\small Causa ltrigger}: {\small\tt induced}, \textbf{\small Cause-phrase}: {\small\tt MMuLV \underline{infection} of non-transgenic mice}, \textbf{\small Effect-phrase}: {\small \tt primarily mature T cell \underline{lymphomas}}$\rangle$


Here, the headwords of the cause and effect phrases are underlined. Intuitively, the headword of a phrase is its most important word and grammatically, it is the ancestor of all the words in a phrase in the sentence's dependency parse tree\footnote{Throughout the paper, we have used dependency relation types as per SpaCy. For detailed explanation of each dependency type, please refer to: \url{https://github.com/clir/clearnlp-guidelines/blob/master/md/specifications/dependency_labels.md}}. Identification of the headword of a cause/effect phrase is an important step in our proposed approach.

\subsection{Proposed Approach}
We propose an algorithm for extracting CE relation triplets which works in following phases:
\subsubsection{Causal Trigger Identification}
In this first phase, a set of causal triggers (words or multi-word expressions) are identified in a given sentence. We observed that the causal triggers can be domain-agnostic (e.g., {\small\tt due to}, {\small\tt because}, {\small\tt caused}) or domain-specific (e.g., {\small\tt inhibits}, {\small\tt down-regulated}). For domain-agnostic causal triggers, we used a list proposed by Girju~\shortcite{girju2003automatic}. Since ensuring the correctness and completeness of causal triggers for the biomedical domain requires a lot of domain knowledge, it was not feasible for us to manually compile such a list. 
Instead, we employed an unsupervised technique for automatically discovering the domain-specific causal verbs as described in~\cite{sharma2018unsupervised}. This technique creates a large list of causal verbs in biomedical domain, using only unlabelled domain corpus and does not require any manual supervision. However, once the list is created, we manually curated it to retain only the high-precision causal verbs. The final list consisted of 109 domain-specific causal triggers and 33 domain-agnostic causal triggers\footnote{Included in the supplementary material}. 
All the morphological variations (e.g., {\small\tt induced}, {\small\tt inducing}) and nominal forms (e.g., {\small\tt induction}) of the causal verbs are also considered. Given any input sentence, this list is looked up for identifying {\em candidate} causal triggers. These are referred to as candidate causal triggers because they become a part of a complete CE triplet only if both of the cause and effect arguments are also identified in the same sentence.

\subsubsection{Cause/Effect Headword Identification} 
In the second phase of our algorithm, for each candidate causal trigger $v$, headwords of its cause and effect argument phrases are identified. All the words in a given sentence which satisfy following conditions are identified as candidate headwords of a cause/effect phrase:

\noindent $\bullet$ All the verbs in the sentence which are not auxiliary of any other main verb, i.e., all the verbs whose dependency relation with their parent is not $aux$.\\
$\bullet$ All the nouns in the sentence which are headwords of any base noun phrase\footnote{A base noun phrase is the noun phrase which does not contain any other noun phrase within it.}, i.e., all the nouns whose dependency relation with their parent is not $compound$.\\
$\bullet$ All other words which play a noun-like role in the dependency parse tree, i.e., the words whose dependency relation with their parents is one of the following: $nsubj$ (nominal subject), $nsubjpass$ (passive nominal subject), $dobj$ (direct object), $pobj$ (prepositional object).
\begin{table*}\small
\begin{tabular}{p{\linewidth}}
\hline
\textbf{Sentence}: {\small\tt Three long-term T-cell \underline{lines}, established from peripheral blood mononuclear cell cultures from three STLV-1-seropositive monkeys, \underline{produced} HTLV-1 Gag and Env antigens and retroviral particles.}\\
\hline
\textbf{Candidate causal trigger:} $v$ = {\small\tt produced}; 
\textbf{Candidate cause/effect headword:} $u$ = {\small\tt lines}\\
\hline
\textbf{Lexical features:}\\
$\bullet$ Actual word tokens corresponding to $v$ and $u$: {\em v.text.}{\small\tt produced}, {\em u.text.}{\small\tt lines}\\
$\bullet$ Rootwords (lemmas) of words corresponding to $v$ and $u$: {\em v.rootword.}{\small\tt produce}, {\em u.rootword.}{\small\tt line}\\
\hline
\textbf{POS tag-based features:}\\
$\bullet$ Part-of-speech (POS) tags of the words corresponding to $v$ and $u$: {\em v.POS.VBD}, {\em u.POS.NNS}\\
$\bullet$ Generalized POS tags of the words corresponding to $v$ and $u$: {\em v.POS\_gen.VERB}, {\em u.POS\_gen.NOUN}\\
\hline
\textbf{Dependency based features:}\\
$\bullet$ Parents/governors in the dependency parse tree for $v$ and $u$: {\em v.parent.text.root}, {\em u.parent.text.}{\small\tt produced}\\
$\bullet$ Dependency relation with parent in the dependency parse tree for $v$ and $u$: {\em v.parent.dep.root}, {\em u.parent.dep.nsubj}\\
$\bullet$ Whether $v$ is an ancestor of $u$ in the dependency parse tree: {\em ancestor.v.u}\\
$\bullet$ Rootword of the ``Lowest Common Ancestor'' (LCA) of $u$ and $v$ in the dependency parse tree: {\em LCA.root\_word.}{\small\tt produce}\\
$\bullet$ Complete path of dependency relations from $u$ to $v$ in the sentence's dependency parse tree: {\em dep.path.u$<$nsubj$<$v}\\ 
{\small (Note: ``{\em $<$DR$<$}'' denotes an edge in the dependency tree labelled with dependency relation {\em DR} where the child is on the left and the parent is on the right. Similarly, ``{\em $>$DR$>$}'' denotes an edge where parent on the left and the child on the right.)}\\
$\bullet$ Whether there is a direct edge between $u$ and $v$ in the dependency parse tree: {\em edge.v.u.nsubj}\\
$\bullet$ Whether any particular dependency relation type lies on the dependency path connecting $u$ to $v$: {\em path.v.u.nsubj}\\
$\bullet$ Whether any particular word lies on the dependency path connecting $u$ to $v$: NA (because, here $u$ is directly connected to $v$)\\
\hline
\textbf{Overall set of features associated with} $\langle v$={\small\tt produced}, $u$={\small\tt lines}$\rangle$ = \{{\em v.text.}{\small\tt produced}, {\em u.text.}{\small\tt lines}, {\em v.rootword.}{\small\tt produce}, {\em u.rootword.}{\small\tt line}, {\em v.POS.VBD}, {\em u.POS.NNS}, {\em v.POS\_gen.VERB}, {\em u.POS\_gen.NOUN}, {\em v.parent.text.root}, {\em u.parent.text.}{\small\tt produced}, {\em v.parent.dep.root}, {\em u.parent.dep.nsubj}, {\em ancestor.v.u}, {\em LCA.rootword.}{\small\tt produce}, {\em dep.path.u$<$nsubj$<$v}, {\em edge.v.u.nsubj}, {\em path.v.u.nsubj}\}\\
\hline
\end{tabular}
\caption{\label{tabFeatures}Various features generated for a pair of a candidate causal trigger and a candidate cause/effect headword}
\end{table*}

For each pair of a causal trigger $v$ and a candidate cause/effect headword $u$ in a given sentence, various types of features are generated. Table~\ref{tabFeatures} describes various types of features with the help of an example $\langle v,u\rangle$ pair. These features are designed to capture various lexical and syntactic characteristics about how any causal trigger $v$ and its corresponding cause/effect argument headword $u$ are mentioned in a given sentence. Then these pairs are classified such that each pair is labelled with any one of the following classes: (i) {\small\sf CAUSE} (indicating that $u$ is the headword of the ``cause'' argument of the causal trigger $v$), (ii) {\small\sf EFFECT} (indicating that $u$ is the headword of the ``effect'' argument of the causal trigger $v$), and (iii) {\small\sf OTHER} (indicating that $u$ is not a cause/effect argument of the causal trigger $v$). If manually annotated $\langle v,u\rangle$ pairs are available, then any supervised classifier can be trained for identifying the above classes, using the features described in Table~\ref{tabFeatures}. However, our goal was to build a system which requires little or no supervision because creation of such an annotated dataset is quite time and effort intensive. Hence, using the same set of features, we employed a decision list algorithm where the rules in the decision list are designed manually. 
Each rule consists of 3 sets of features (described in Table~\ref{tabFeatures}) as follows:

\noindent $\bullet$ \textbf{AND set}: This set should be non-empty and each feature in this set should be present in the features set associated with any $\langle v,u\rangle$ pair for the rule to be satisfied. Here, $v$ is a causal trigger and $u$ is a candidate headword of a cause/effect argument.\\
$\bullet$ \textbf{OR set}: This set may be empty but if it is non-empty, then at least one feature from this set should be present in the features set associated with any $\langle v,u\rangle$ pair for the rule to be satisfied.\\
$\bullet$ \textbf{NEG set}: This set may be empty but if it is non-empty, then none of features from this set should be present in the features set associated with any $\langle v,u\rangle$ pair for the rule to be satisfied.

In other words, each rule must specify a conjunction of some features. Optionally, it may specify other sets of features in addition to the conjunction, which behave as a disjunction or negation of some other features. E.g., consider the following rules:

\noindent $\bullet$ One of the rules to identify CAUSE is--\\
\textbf{AND}: \{{\em dep.path.u$<$nsubj$<$v}\}; \newline \textbf{OR}: \{{\em u.POS\_gen.NOUN}, {\em u.POS\_gen.PROPN}\}; \newline \textbf{NEG}: \{{\em v.rootword.}{\small\tt result}\}\\
This rule will be satisfied for a $\langle v,u\rangle$ pair if the the dependency path from the candidate headword $u$ to the causal trigger $v$ is {\em u$<$nsubj$<$v} (i.e., $u$ is a nominal subject of $v$) . In addition, the generalized POS tag of the candidate headword $u$ should be either {\em NOUN} (common noun) or {\em PROPN} (proper noun). Also, the trigger $v$ should not be any morphological variation of {\small\tt result}.
\\
$\bullet$ Another rule to identify CAUSE is--\\
\textbf{AND}: \{{\em v.POS\_gen.VERB}, {\em dep.path.u$>$relcl$>$v}, {\em v.child.}{\small\tt which}\}; \textbf{OR}: \{ \};\\ \textbf{NEG}: \{u.parent.dep.attr\}\\
This rule will be satisfied for a $\langle v,u\rangle$ pair if the causal trigger $v$ is a verb, the dependency path from $u$ to $v$ is {\em u$>$relcl$>$v} (i.e. a relative clause headed at $v$ modifies $u$), and {\small\tt which} is one of the children of $v$. Moreover, the dependency relation of $u$ with its parent should NOT be {\em attr}. There are no disjunctive features for this rule.

Our decision list has $33$ and $21$ rules for identifying {\small\sf CAUSE} and {\small\sf EFFECT} headwords, respectively. Tables~\ref{tabCauseRules} and~\ref{tabEffectRules} show the most important rules for identifying {\small\sf CAUSE} and {\small\sf EFFECT} headwords, respectively\footnote{All rules are included in the supplementary material}. Table~\ref{tabRulesApplication} illustrates the application of these rules for an example sentence.

\begin{table*}\small
\begin{tabular}{p{\linewidth}}
\hline
{\small\tt PEDV belongs to the Alphacoronavirus genus and can cause a highly contagious enteric disease.}\\
\hline
\includegraphics[width=0.95\linewidth, height=0.25\linewidth]{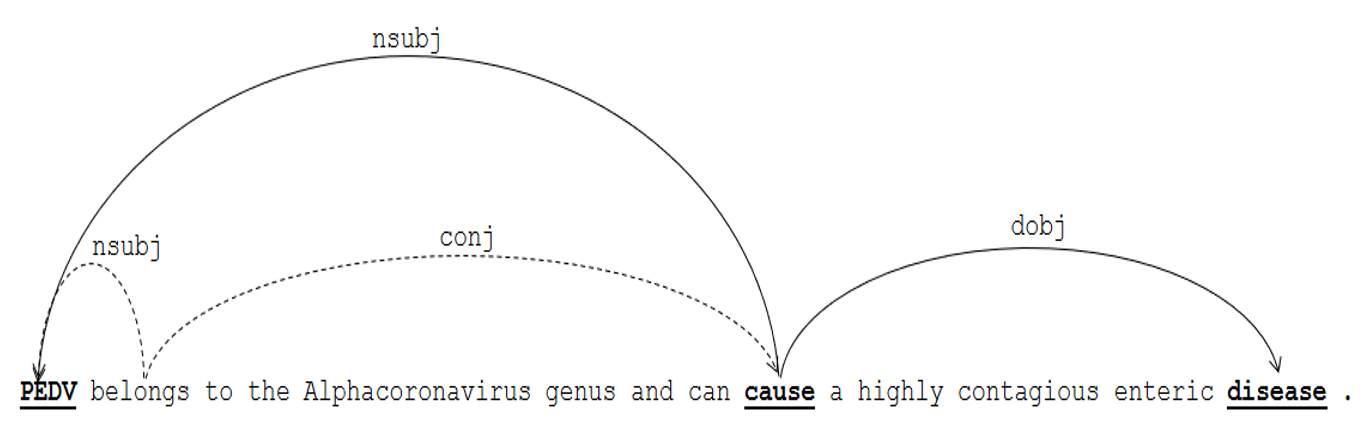}\\
\hline
$\langle v=${\small\tt cause}, $u=${\small\tt PEDV}$\rangle$: {\small\sf CAUSE} 
(Rule ID 1 in Table~\ref{tabCauseRules})\\
$\langle v=${\small\tt cause}, $u=${\small\tt disease}$\rangle$: {\small\sf EFFECT} 
(Rule ID 1 in Table~\ref{tabEffectRules})\\
\hline
\end{tabular}
\caption{\label{tabRulesApplication}An illustration of application of rules to identify headword described in Tables~\ref{tabCauseRules} and~\ref{tabEffectRules}}
\end{table*}

\subsubsection{Cause/Effect Phrase Expansion}
In the third phase of our algorithm, the cause/effect headwords identified in the second phase are expanded to get the complete phrases. Here, we again use dependency tree and simply consider the complete subtree rooted under a given headword. E.g., in Table~\ref{tabRulesApplication}, the headword {\small\tt disease} is expanded using the subtree rooted at {\small\tt disease} which is {\small\tt a highly contagious enteric disease}. We also use a simple rules specification language
for specifying some exceptions for phrase expansion, as follows: 

\noindent $\bullet$ An option to exclude children/dependants having certain dependency relations with their parents. 
In our experiments, we exclude following dependency relations -- $punct$ (connects to punctuation symbols), $appos$ (connects to an appositive phrase), $advcl$ (connects to an adverbial clause)\\
$\bullet$ An option to limit the phrase boundaries to the left/right of the trigger word. If the trigger word itself is a descendent of a cause/effect headword, then the cause/effect phrase boundaries will never include the trigger word and exceed beyond it.


\subsubsection{Cause-Effect Triplet Formation}
In this final phase, CE triplets are formed for each trigger $v$. 
Let $U_C$ be the set of {\small\sf CAUSE} headwords identified for $v$ and $U_E$ be the set of {\small\sf EFFECT} headwords identified for $v$. Then, the final set of CE triplets associated with the trigger $v$ is:\\
\{$\langle u_1, v, u_2$ $\rangle |$ s.t. $(u_1, u_2) \in U_C\times U_E$\}.

Thus, the final CE triplet for the sentence in Table~\ref{tabRulesApplication} will be: $\langle${\small\tt PEDV}, {\small\tt cause}, {\small\tt a highly contagious enteric disease} $\rangle$.

\subsection{Extraction of Additional Arguments}
In addition to cause and effect arguments, we extract two more arguments of a cause-effect relation: i) negation, and ii) uncertainty. If the causal trigger has a child in its dependency tree with dependency relation $neg$, then we extract it as a {\em negation} argument. E.g., {\small\tt [Overnight \underline{incubation} with 1 microM safrole]$_{Cause}$ did \textbf{[\underline{not}]}$_{Negation}$ [\underline{alter}]$_{Trigger}$ [cell \underline{proliferation}]$_{Effect}$.} Here, the causal trigger is {\small\tt alter} is negated by {\small\tt not} which is extracted as a {\em negation} argument.

Similarly, if the causal trigger has a child in its dependency tree with dependency relation $aux$ and it is from a set of uncertainty indicating words (such as {\small\tt may}, {\small\tt might}, {\small\tt would}), then we extract it as an {\em uncertainty} argument. E.g., {\small\tt [\underline{Glucocorticoids}]$_{Cause}$ \textbf{[\underline{might}]}$_{Uncertainty}$ [\underline{induce}]$_{Trigger}$ [the \underline{apoptosis} of some types of AML cells]$_{Effect}$, just like that of some lymphoid leukemia cells.} Here, the causal trigger is {\small\tt induce} is modified by {\small\tt might} which is extracted as an {\em uncertainty} argument.

\section{Related Work}
Extensive research is going on in the field of text mining in the biomedical domain. 
Several interesting approaches have been proposed to extract named entities, relationships, summaries, text classification, ontologies, knowledge discovery graphs and hypothesis generation~\cite{luque2019advanced}. 

A lot of approaches for biomedical relation extraction use some form of Named Entity Recognition(NER) to identify the medical concepts in a sentence first and then find relations between them using rule based techniques, machine learning algorithms or a combination of both~\cite{rink2011automatic,uzuner2010semantic,zhu2013detecting,xu2015large}. Identifying named entities in the first step helps to reduce the number of candidates for relation identification. However, the extraction using this technique suffers if the medical concept is not picked up or identified incorrectly by the NER module. Our technique is different from these techniques as we do not rely on NER for our CE triplet extraction. As a result we are able to extract CE triplets even in those cases when a named entity can be missed by the NER system. The most similar line of research to our work is SemRep and its corresponding knowledge base SemMedDB.

\subsection{SemRep and SemMedDB}
SemRep~\cite{rindflesch2003interaction} is a UMLS-based\footnote{\url{https://www.nlm.nih.gov/research/umls/index.html}} program that extracts three-part semantic predications, from sentences in biomedical text. These predications are in the form:
\begin{center}
$<$subject, RELATION, object$>$
\end{center}
The subject and object arguments are UMLS Metathesaurus Concepts~\cite{bodenreider2004unified}, while the relation between them is from the UMLS Semantic Network. SemRep works by indentifying UMLS Concepts present in a sentence and then using rules to determine which UMLS relation exists between them. 
The Semantic MEDLINE Database (SemMedDB)~\cite{kilicoglu2012semmeddb} is a repository of predications extracted from all of PubMed citations using SemRep. 
SemRep extracts relations of various types but only a few of them represent causal relations, which are: {\small\sf AFFECTS}, {\small\sf CAUSES}, {\small\sf STIMULATES}, {\small\sf INHIBITS}, {\small\sf DISRUPTS}, {\small\sf PRODUCES}, {\small\sf PRECEDES}, {\small\sf COMPLICATES}, {\small\sf PREDISPOSES}, {\small\sf PREVENTS}. Although our approach does not explicitly identify relation types, it extracts all types of Cause-Effect relations, and the finer interpretation of each relation can be inferred from the corresponding causal trigger.

\section{Experimental Analysis}
We evaluate our proposed approach for cause-effect relation extraction along multiple aspects.

\subsection{Leukaemia-related PubMed Abstracts Corpus}
We created a corpus of 58,761 PubMed citations (title as well as abstract) which are related to Leukaemia. We pre-processed this dataset using SpaCy~\cite{spacy2} to obtain -- tokens, sentences, and dependency parse trees for each sentence. Overall, the dataset consisted of 568,528 sentences, i.e. on an average each citation contains 9.67 sentences. Also, the median sentence length is 27 words, indicating that the sentences in the corpus are fairly complex.

We applied our Cause-Effect relation extraction approach on this dataset to extract Cause-Effect (CE) triplets. Overall, 152,655 CE triplets were extracted. As there are no gold-standard annotations available for this data, we use random sampling to estimate precision. The detailed evaluation methods are described in following sections.

\begin{table*}\small
\begin{tabular}{p{0.61\linewidth}ccp{0.14\linewidth}}
\hline
\multirow{2}{*}{\textbf{Sentence with a $\langle v,u \rangle$ pair marked inline}} & \textbf{RuleID} & \multirow{2}{*}{\textbf{Score}} & \multirow{2}{*}{\textbf{Comment}} \\
 & \textbf{(C/E)} &  &  \\
\hline
{\small\tt In particular, we reported the existence of BCR-ABL alternative splicing isoforms, in about 80\% of Philadelphia-positive \textbf{[\underline{patients}]}$_u$, which \textbf{[\underline{lead}]}$_v$ to the expression of aberrant proteins.} & 8 ({\small\sf CAUSE}) & 0 & Incorrect (due to incorrect parsing)\\
\hline
{\small\tt Childhood acute myeloid leukemia with bone marrow eosinophilia \textbf{[\underline{caused}]}$_v$ by \textbf{[\underline{t(16}]}$_u$ ; 21)(q24 ; q22).} & 3 ({\small\sf CAUSE}) & 1 & Partially correct (due to incorrect tokenization) \\
\hline
{\small\tt Perifosine and TRAIL synergized to activate caspase-8 and induce apoptosis, which was \textbf{[\underline{blocked}]}$_v$ by a caspase-8-selective \textbf{[\underline{inhibitor}]}$_u$.} & 3 ({\small\sf CAUSE}) & 2 & Correct \\
\hline
{\small\tt Monocytic \textbf{[\underline{maturation}]}$_u$ ( morphologic and immunologic ) was \textbf{[\underline{induced}]}$_v$ in all cases studied , although to different rates, by TNF-alpha and by HTR-9 incubation.} & 3 ({\small\sf EFFECT}) & 2 & Correct \\
\hline
\end{tabular}
\caption{\label{tabScoresHeadwordRules}Examples of scores assigned by the human expert for evaluating headword identification rules. The candidate trigger $v$ and the candidate headword $u$ are underlined and marked within the sentence.}
\end{table*}

\subsubsection{Evaluation of Cause/Effect Headword Identification Rules} 
We estimate the precision of each linguistic rule used in our decision list to identify cause/effect headwords. Our decision list is used to classify a pair of a causal trigger and a candidate headword of a cause/effect phrase into 3 different classes -- {\small\sf CAUSE}, {\small\sf EFFECT} and {\small\sf OTHER}. As described earlier, there are 33 rules to identify {\small\sf CAUSE} and 21 rules to identify {\small\sf EFFECT}. To estimate precision of a {\small\sf CAUSE}-predicting rule, we randomly select 10 CE triplets from the 152,655 CE triplets such that the cause headword in those triplets was identified using that particular rule. We also ensured that our random sample contains equal mixture of simple and complex sentences. Out of the 10 CE triplets, we ensure that 5 of them are extracted from a sentence whose length is more than 27 words (median length) and the rest 5 are shorter than 27 words. A human expert then evaluated these triplets manually to assign a score to each triplet. The scale used to evaluate is -- 0 if completely incorrect, 1 if partially correct and 2 if completely correct (see Table~\ref{tabScoresHeadwordRules} for examples). The precision for the rule is computed by dividing the total score for 10 CE triplets by 20. Table~\ref{tabCauseRules} shows precision as well as {\em coverage} for 10 rules for identifying {\small\sf CAUSE} headword. Similarly, Table~\ref{tabEffectRules} shows precision as well as coverage for 10 rules for identifying {\small\sf EFFECT} headword. In both the cases, the tables show only top 10 rules as per their coverage. Here, the coverage is the total number of extracted CE triplets for which a certain rule was used for identifying a {\small\sf CAUSE} or {\small\sf EFFECT} headword.

\begin{table*}[t]\small
\begin{tabular}{cp{0.57\linewidth}ccc}
\hline
\textbf{Rule} & \multirow{2}{*}{\textbf{Rule}} & \multirow{2}{*}{\textbf{Coverage}} & \multirow{2}{*}{\textbf{Prec.}} & \textbf{\small \#CE Triplets}\\
\textbf{ID} &  &  &  & \textbf{\small $\notin$ SMDB-L} \\
\hline
1 & AND: \{{\em dep.path.u$<$nsubj$<$v}\}; \newline OR: \{{\em u.POS\_gen.NOUN}, {\em u.POS\_gen.PROPN}\}; \newline NEG: \{{\em v.rootword.}{\small\tt result}\} & 69041 (45.2\%) & 0.8 & 21104 \\
\hline
2 & AND: \{{\em dep.path.u$<$npadvmod$<$v}\} & 20263 (13.3\%) & 0.8 & 6128 \\
\hline
3 & AND: \{{\em v.POS\_gen.VERB}, {\em dep.path.u$<$pobj$<$agent$<$v}\} & 15113 (9.9\%) & 0.95 & 4123 \\
\hline
4 & AND: \{{\em v.POS\_gen.VERB}, {\em dep.path.u$>$relcl$>$v}, {\em v.child.}{\small\tt that}\} & 5534 (3.6\%) & 0.85 & 1564 \\
\hline
5 & AND: \{{\em v.text.}{\small\tt due}, {\em dep.path.u$<$pobj$<$v}\} & 4012 (2.6\%) & 0.7 & 1252 \\
\hline
6 & AND: \{{\em v.POS\_gen.VERB}, {\em dep.path.u$<$nsubj$<$LCA$>$prep$>$pcomp$>$v}\} & 3157 (2.1\%) & 0.6 & 931 \\
\hline
7 & AND: \{{\em v.POS\_gen.NOUN}, {\em dep.path.u$<$pobj$<$prep$<$v}, {\em path.}{\small\tt by}\} & 2900 (1.9\%) & 0.95 & 845 \\
\hline
8 & AND: \{{\em v.POS\_gen.VERB}, {\em dep.path.u$>$relcl$>$v}, {\em v.child.}{\small\tt which}\}; \newline NEG: \{{\em u.parent.dep.attr}\} & 2889 (1.9\%) & 0.6 & 852 \\
\hline
9 & AND: \{{\em v.POS\_gen.VERB}, {\em dep.path.u$<$nsubjpass$<$LCA$>$xcomp$>$v}\} & 2836 (1.8\%) & 1.0 & 880 \\
\hline
10 & AND: \{{\em v.rootword.}{\small\tt role}, {\em dep.path.u$<$nsubj$<$LCA$>$dobj$>$v}, \newline {\em lca.rootword.}{\small\tt play}\} & 2679 (1.7\%) & 0.9 & 551 \\
\hline
%
%
%
\end{tabular}
\caption{\label{tabCauseRules}Performance of the rules identifying {\small\sf CAUSE} headwords ($v$ is a candidate causal trigger and $u$ is a candidate headword of a cause phrase)}
\end{table*}

\begin{table*}[t]\small
\begin{tabular}{cp{0.57\linewidth}ccc}
\hline
\textbf{Rule} & \multirow{2}{*}{\textbf{Rule}} & \multirow{2}{*}{\textbf{Coverage}} & \multirow{2}{*}{\textbf{Prec.}} & \textbf{\small \#CE Triplets}\\
\textbf{ID} &  &  &  & \textbf{\small $\notin$ SMDB-L} \\
\hline
1 & AND: \{{\em edge.v.u.dobj}\} & 84072 (55.1\%) & 0.8 & 25783 \\
\hline
2 & AND: \{{\em v.POS\_gen.VERB}, {\em dep.path.u$>$amod$>$v}\} & 18972 (12.4\%) & 1.0 & 5668 \\ 
\hline
3 & AND: \{{\em edge.v.u.nsubjpass}\} & 9785 (6.4\%) & 0.9 & 2692 \\
\hline
4 & AND: \{{\em v.POS.VBN}, {\em edge.u.v.acl}, {\em u.POS\_gen.NOUN}, {\em v.POS\_gen.VERB}\} &  6300 (4.1\%) & 1.0 & 1750 \\
\hline
5 & AND: \{{\em dep.path.u$>$prep$>$v}\}; OR:	\{{\em v.text.}{\small\tt because}, {\em v.text.}{\small\tt due}\} & 6223 (4.1\%) & 0.8 & 1996 \\
\hline
6 & AND: \{{\em v.POS\_gen.VERB}, {\em v.rootword.}{\small\tt lead}, {\em dep.path.u$<$pobj$<$prep$<$v}, {\em path.}{\small\tt to}\} & 4897 (3.2\%) & 1.0 & 1441 \\
\hline
7 & AND: \{{\em v.POS\_gen.NOUN}, {\em dep.path.u$<$pobj$<$prep$<$v}, {\em path.of}\};\newline OR: \{{\em v.rootword.}{\small\tt cause}, {\em v.rootword.}{\small\tt reason}, {\em v.child.prep.}{\small\tt by}, \newline{\em v.child.agent.}{\small\tt by}\} & 3850 (2.5\%) & 1.0 & 1234 \\
\hline
8 & AND: \{{\em v.POS\_gen.VERB}, {\em v.rootword.}{\small\tt contribute}, \newline{\em dep.path.u$<$pobj$<$prep$<$v}, {\em path.}{\small\tt to}\} & 3438 (2.3\%) & 1.0 & 933 \\
\hline
9 & AND: \{{\em v.rootword.}{\small\tt role}, {\em dep.path.u$<$pobj$<$prep$<$v}, {\em path.}{\small\tt in}\} & 2365 (1.5\%) & 1.0 & 463 \\
\hline
10 & AND: \{{\em dep.path.u$>$advcl$>$mark$>$v}, {\em v.text.}{\small\tt because}\} & 2128 (1.4\%) & 0.8 & 853 \\
\hline
\end{tabular}
\caption{\label{tabEffectRules}Performance of the rules identifying {\small\sf EFFECT} headwords ($v$ is a candidate causal trigger and $u$ is a candidate headword of an effect phrase)}
\end{table*}

\subsubsection{Evaluation of Phrase Expansion Rules}
In order to evaluate the performance of the phrase expansion step, the human expert also evaluated correctness of phrases along with annotations obtained for headword identification rules (10 random CE triplets for each rule as explained earlier). The same scale of 0 to 2 was used here as was used in case of headword identification.  
The phrase expansion accuracy was observed to be 95.29\% and 94.27\% for {\small\sf CAUSE} and {\small\sf EFFECT} phrases, respectively. For computing this accuracy, only those phrases are considered whose headwords were identified correctly.

\subsubsection{Comparison with SemMedDB}
We applied our proposed approach on this corpus of 58,761 Leukaemia-related PubMed citations and obtained 152,655 cause-effect triplets. 
For comparison, we also considered a subset of SemMedDB for the same set of 58,761 PubMed citations, which we refer to as \textbf{\em SMDB-L}. Out of 503,183 predications in {\em SMDB-L}, only 77,135 correspond to the causal predicates.
\begin{table*}\small
\begin{tabular}{p{0.73\linewidth}cp{0.14\linewidth}}
\hline
\textbf{Sentence with a CE triplet marked inline} & \textbf{Score} & \textbf{Comment}\\
\hline
{\small\tt Furthermore, \textbf{[the current diagnostic \underline{interpretation} of flow cytometry readouts]}$_{Effect}$ is \textbf{[influenced]}$_{Trigger}$ arbitrarily by individual experience and \textbf{[\underline{knowledge}]}$_{Cause}$.} & 0 & Incorrect (not causal) \\
\hline
{\small\tt \textbf{[CsA \underline{treatment}]}$_{Cause}$ \textbf{[resulted]}$_{Trigger}$ in \textbf{[an increased \underline{incidence} of hyperbilrubinemia , which rapidly reversed upon conclusion of drug therapy]}$_{Effect}$ .} & 1 & Partially correct (due to long effect phrase) \\
\hline
{\small\tt We conclude that \textbf{[\underline{GM-CSF}]}$_{Cause}$ is effective in improving CLL associated chronic neutropenia and also \textbf{[enhances]}$_{Trigger}$ \textbf{[impaired granulocyte \underline{chemiluminescence}]}$_{Effect}$.} & 2 & Correct \\
\hline
\end{tabular}
\caption{\label{tabScoresCETriplets}Examples of scores assigned by the human expert for evaluating CE triplets. The trigger and cause/effect phrases are marked inline in the sentences. The headwords of cause/effect phrases are underlined. Note that each row represents one CE triplet and there may be more CE triplets extracted in the same sentence.}
\end{table*}
\begin{table}[b]\small\center
\begin{tabular}{p{0.4\linewidth}p{0.2\linewidth}p{0.2\linewidth}}
\hline
\textbf{Type of CE Triplets} & \textbf{Precision (Strict)} & \textbf{Precision (Lenient)} \\
\hline
CE Triplets $\notin$ SMDB-L & 0.60 & 0.74 \\
\hline
\end{tabular}
\caption{\label{tabComparisonSMDBL}Precision of extracted CE triplets which is estimated by manually evaluating random 100 triplets}
\end{table}

Our proposed approach is able to extract almost twice the number of CE triplets as compared to {\em SMDB-L} (152,655 vs 77,135). We estimate the precision of our CE triplets which are extracted over an above {\em SMDB-L} using a random sample. We randomly select 100 CE triplets which are extracted from those sentences for which {\em SMDB-L} does not any predication having a causal predicate. Here, the causal predicates are -- {\small\sf AFFECTS}, {\small\sf CAUSES}, {\small\sf STIMULATES}, {\small\sf INHIBITS}, {\small\sf DISRUPTS}, {\small\sf PRODUCES}, {\small\sf PRECEDES}, {\small\sf COMPLICATES}, {\small\sf PREDISPOSES}, {\small\sf PREVENTS}. Examples of non-causal predicates are {\small\sf PART\_OF}, {\small\sf TREATS}, {\small\sf PROCESS\_OF}. A human expert then evaluated these randomly selected 100 CE triplets and assigned a score to each triplet. The same scale used earlier to evaluate headword identification rules is used, i.e., 0 if completely incorrect, 1 if partially correct and 2 if completely correct (see examples in Table~\ref{tabScoresCETriplets}). We compute two precision values: i) strict precision is computed by considering partially correct to be incorrect, and ii) lenient precision is computed by considering the sum of scores of all triplets divided by 200. Table~\ref{tabComparisonSMDBL} shows both of these precision values for the triplets extracted over and above {\em SMDB-L}. Moreover, the last column in Tables~\ref{tabCauseRules} and~\ref{tabEffectRules} show the number of CE triplets extracted by each rule which are not part of {\em SMDB-L}. These CE triplets are extracted from sentences for which {\em SMDB-L} does have not any predication with a causal predicate.


\subsection{Gold-standard Dataset}
We obtained from SemRep website\footnote{\url{https://semrep.nlm.nih.gov/GoldStandard.html}}, a gold-standard dataset~\cite{kilicoglu2011constructing} for semantic predications where subject/object arguments are manually annotated. This dataset contains 500 sentences annotated with 1371 semantic predications annotated by human experts. Out of these 1371 predications, only 258 correspond to causal predicates (the same list of predicates considered in {\em SMDB-L}). 
We ignored the non-causal predications as the scope of this work is limited to identifying only causal relations. 

Each predication in this dataset consists of following important fields:

\noindent $\bullet$ Predicate/Relation type\\
$\bullet$ Subject name, its concept ID from ULMS Metathesaurus, and corresponding text span which is analogous to {\em Cause} phrase in our CE triplets\\
$\bullet$ Object name, its concept ID from UMLS Metathesaurus, and corresponding text span which is analogous to {\em Effect} phrase in our CE triplets
\begin{table}\small\center
\begin{tabular}{cccc}
\hline
\textbf{Approach} & \textbf{Precision} & \textbf{Recall} & \textbf{F1-measure} \\
\hline
SemRep & 58.78 & 29.84 & 39.59 \\
Proposed & 50.83 & 35.66 & \textbf{41.91} \\
\hline
\end{tabular}
\caption{\label{tabResultsGoldStandard}Performance of our proposed approach for Cause-Effect relation extraction as compared to SemRep over the gold-standard dataset.}
\end{table}

\subsubsection{Baseline: SemRep}
In order to compare the performance of our proposed approach on the gold-standard dataset with SemRep, we processed its 500 sentences using the online SemRep Batch Facility\footnote{\url{https://ii.nlm.nih.gov/Batch/UTS_Required/semrep.shtml}}. We set the {\em Knowledge Source} and the {\em Lexicon Year} as {\em 2018} and selected the {\em Strict Model}. 
We obtained the {\em Full Fielded Model Output} because it provided the text spans of the subject and object in the sentence. These text spans of subject and object correspond to cause and effect phrases in our CE triplet format, respectively. 

\subsubsection{Evaluation}
The format of our CE triplet is $\langle${\em cause phrase}, {\em trigger}, {\em effect phrase}$\rangle$ whereas each predication in the gold-standard dataset and SemRep output is considered in the form - $\langle${\em subject text}, {\em predicate}, {\em object text}$\rangle$. As we are considering only the causal predicates, we ignore the actual trigger/predicate type while evaluating. Hence, a CE triplet is considered to be {\em matching} a gold-standard predication if there is an overlap of at least one content word between the {\em cause phrase} and the {\em subject text} as well as between the {\em effect phrase} and the {\em object text}. E.g., consider a gold-standard predication $\langle$ {\small\tt Calcitonin gene-related peptide}, {\small\sf INHIBITS}, {\small\tt monocyte chemoattractant protein-1} $\rangle$. We consider it to be matching with our predicted CE triplet $\langle${\small\tt Calcitonin gene-related peptide}, {\small\tt inhibits}, {\small\tt interleukin-1beta-induced endogenous monocyte chemoattractant protein-1 secretion in type II alveolar epithelial cells}$\rangle$.

Similarly, a SemRep predication is considered to be {\em matching} a gold-standard predication if there is an overlap of at least one content word between the corresponding {\em subject texts} as well as between the corresponding {\em object texts}. For each gold-standard predication, a true positive (TP) is counted for our proposed approach if there is a {\em matching} CE triplet and a false negative (FN) is counted otherwise. Also, for each predicted CE triplet, a false positive (FP) is counted if there is no {\em matching} predication in the gold-standard dataset. Precision, Recall and F1-measure are then computed as follows. 
\begin{equation}\small
P=\frac{TP}{TP+FP}, R=\frac{TP}{TP+FN}, F1=\frac{2\cdot P\cdot R}{P+R}
\end{equation}
For the SemRep baseline, the evaluation is carried out in a similar manner. Table~\ref{tabResultsGoldStandard} shows the performance of our proposed technique as compared with the SemRep baseline. 

We analyzed our extracted CE triplets for low precision. One of the reason is that unlike SemRep, our technique does not restrict cause/effect phrases to be of certain ``semantic types''. Hence, we have several False Positive extractions where cause/effect phrases may not be normalized to any UMLS concept. E.g., consider the sentence: {\small\tt IFN-alpha profoundly alters cytoskeletal organization of hairy cells and causes reversion of the hairy appearance into a rounded morphology.} Here, we extract the following CE triplet which is correct but not annotated in the gold-standard dataset and hence is counted as a False Positive: $\langle$ {\small\tt IFN-alpha}, {\small\tt causes}, {\small\tt reversion of the hairy appearance into a rounded morphology} $\rangle$. 

\section{Conclusion and Future Work}
We proposed a knowledge-based Cause-Effect relation extraction approach which does not require any kind of supervision or large annotated corpus. The proposed approach is based on: i) an unsupervised machine learning technique to discover causal triggers, and ii) a set of high-precision linguistic rules to identify cause/effect arguments of these causal triggers. We evaluated our approach using a large corpus of 58,761 Leukaemia-related PubMed abstracts consisting of 568,528 sentences. We extracted 152,655 CE triplets from this corpus where each triplet consists of a cause phrase, an effect phrase and a causal trigger. As compared to the existing knowledge base -- SemMedDB~\cite{kilicoglu2012semmeddb}, the number of extractions is almost twice, i.e., for the same set of PubMed abstracts, SemMedDB has only 77,135 predications corresponding to causal predicates. In addition, we also evaluated our approach on a gold-standard annotated corpus of 500 sentences and it outperformed the existing technique SemRep~\cite{rindflesch2003interaction} on this corpus. 
In future, we plan to extend this work in several aspects:\\
$\bullet$ Sentence simplification: Upon analysis of errors, we observed that a large fraction of errors are occurring due to incorrect dependency parsing. Hence, even if our linguistic rules are correct, due to incorrect dependency paths we get wrong extractions. A major reason behind this is that the sentences in biomedical domain are more complex (median length of 27 words) than the general domain sentences (on which dependency parsers are generally trained). Hence, it would be interesting to simplify sentences before parsing using any sentence simplification tool.\\
$\bullet$ Distant supervision: Even though our linguistic rules are high-precision, it is difficult to scale-up the recall and maintain these rules. Hence, we plan to use these rules to automatically create a large annotated dataset of Cause-Effect relations with possibly noisy annotations and then train a supervised machine learning model using this dataset.\\
$\bullet$ Use of sophisticated knowledge validation techniques to automatically improve the quality of extracted CE triplets.

\bibliography{anthology,ref}
\bibliographystyle{acl_natbib}

\section*{Supplementary Material}
Table~\ref{tabTriggers} shows the complete list of domain-agnostic and domain-specific causal triggers.


\begin{table*}[h]\small
\begin{tabular}{p{\linewidth}}
\hline
\textbf{Domain-agnostic cue-phrases and causal verbs:}\\
\hline
{\small\tt cause of}, {\small\tt causes of}, {\small\tt cause for}, {\small\tt causes for}, {\small\tt reason for}, {\small\tt reasons for}, {\small\tt reason of}, {\small\tt reasons of}, {\small\tt as a consequence}, {\small\tt as a result}, {\small\tt due}, {\small\tt because}, {\small\tt activate}, {\small\tt bring about}, {\small\tt cause}, {\small\tt contribute to}, {\small\tt create}, {\small\tt derive from}, {\small\tt effect}, {\small\tt elicit}, {\small\tt entail}, {\small\tt evoke}, {\small\tt generate}, {\small\tt give rise to}, {\small\tt implicate in}, {\small\tt lead to}, {\small\tt originate in}, {\small\tt provoke}, {\small\tt result from}, {\small\tt stem from}, {\small\tt stimulate}, {\small\tt trigger off}, {\small\tt role}\\
\hline
\textbf{Domain-specific causal verbs:}\\
\hline
{\small\tt coadministrate}, {\small\tt down-regulate}, {\small\tt up-regulate}, {\small\tt co-express}, {\small\tt re-express}, {\small\tt over-express}, {\small\tt dysregulate}, {\small\tt degranulate}, {\small\tt knockdown}, {\small\tt ablate}, {\small\tt abrogate}, {\small\tt accelerate}, {\small\tt advance}, {\small\tt affect}, {\small\tt alter}, {\small\tt attenuate}, {\small\tt benefit by}, {\small\tt benefit from}, {\small\tt block}, {\small\tt convert}, {\small\tt decrease}, {\small\tt degrade}, {\small\tt delineate}, {\small\tt deplete}, {\small\tt deregulate}, {\small\tt die of}, {\small\tt diminish}, {\small\tt discharge}, {\small\tt disrupt}, {\small\tt disseminate}, {\small\tt divide}, {\small\tt elevate}, {\small\tt eliminate}, {\small\tt enforce}, {\small\tt enhance}, {\small\tt enrich}, {\small\tt eradicate}, {\small\tt exacerbate}, {\small\tt exert}, {\small\tt expand}, {\small\tt extend}, {\small\tt fuse}, {\small\tt govern}, {\small\tt impact}, {\small\tt impair}, {\small\tt improve}, {\small\tt increase}, {\small\tt induce}, {\small\tt infect}, {\small\tt infiltrate}, {\small\tt influence}, {\small\tt inhibit}, {\small\tt inject}, {\small\tt intensify}, {\small\tt kill}, {\small\tt knock down}, {\small\tt maximize}, {\small\tt mediate}, {\small\tt minimize}, {\small\tt optimize}, {\small\tt originate from}, {\small\tt portend}, {\small\tt prevent}, {\small\tt produce}, {\small\tt proliferate}, {\small\tt prolong}, {\small\tt protect}, {\small\tt reactivate}, {\small\tt reduce}, {\small\tt regain}, {\small\tt regulate}, {\small\tt relapse}, {\small\tt remove}, {\small\tt replicate}, {\small\tt repress}, {\small\tt reproduce}, {\small\tt rescue}, {\small\tt restore}, {\small\tt reverse}, {\small\tt revert}, {\small\tt sensitize}, {\small\tt shorten}, {\small\tt stabilize}, {\small\tt substitute}, {\small\tt suppress}, {\small\tt transfer}, {\small\tt transform}, {\small\tt trigger}, {\small\tt transplant}, {\small\tt escalate}, {\small\tt complicate}, {\small\tt express}, {\small\tt progress}, {\small\tt decline}, {\small\tt predispose}, {\small\tt translate}, {\small\tt secrete}, {\small\tt unblock}, {\small\tt grow}, {\small\tt remit}, {\small\tt remove}, {\small\tt abolish}, {\small\tt drive}, {\small\tt modulate}, {\small\tt amplify}, {\small\tt antagonize}, {\small\tt destruct}, {\small\tt destroy}, {\small\tt lower} \\
\hline
\end{tabular}
\caption{\label{tabTriggers}List of cue-phrases and causal verbs used in the first phase (Causal Trigger Identification).}
\end{table*}

\subsection*{Rules for Cause/Effect Headword Identification}
Our decision list has $33$ and $21$ rules for identifying {\small\sf CAUSE} and {\small\sf EFFECT} headwords, respectively. Tables~\ref{tabCauseRules} and~\ref{tabEffectRules} show all these rules for identifying {\small\sf CAUSE} and {\small\sf EFFECT} headwords, respectively.

\begin{table*}[h]\footnotesize
\begin{tabular}{cp{0.9\linewidth}}
\hline
\textbf{Rule} & \textbf{Rule}\\
\hline
1 & AND: \{{\em dep.path.u$<$nsubj$<$v}\}; \newline OR: \{{\em u.POS\_gen.NOUN}, {\em u.POS\_gen.PROPN}\}; \newline NEG: \{{\em v.rootword.}{\small\tt result}\} \\
\hline
2 & AND: \{{\em dep.path.u$<$npadvmod$<$v}\} \\
\hline
3 & AND: \{{\em v.POS\_gen.VERB}, {\em dep.path.u$<$pobj$<$agent$<$v}\}\\
\hline
4 & AND: \{{\em v.POS\_gen.VERB}, {\em dep.path.u$>$relcl$>$v}, {\em v.child.}{\small\tt that}\}\\
\hline
5 & AND: \{{\em v.text.}{\small\tt due}, {\em dep.path.u$<$pobj$<$v}\}\\
\hline
6 & AND: \{{\em v.POS\_gen.VERB}, {\em dep.path.u$<$nsubj$<$LCA$>$prep$>$pcomp$>$v}\}\\
\hline
7 & AND: \{{\em v.POS\_gen.NOUN}, {\em dep.path.u$<$pobj$<$prep$<$v}, {\em path.}{\small\tt by}\}\\
\hline
8 & AND: \{{\em v.POS\_gen.VERB}, {\em dep.path.u$>$relcl$>$v}, {\em v.child.}{\small\tt which}\}; \newline NEG: \{{\em u.parent.dep.attr}\}\\
\hline
9 & AND: \{{\em v.POS\_gen.VERB}, {\em dep.path.u$<$nsubjpass$<$LCA$>$xcomp$>$v}\}\\
\hline
10 & AND: \{{\em v.rootword.}{\small\tt role}, {\em dep.path.u$<$nsubj$<$LCA$>$dobj$>$v}, \newline {\em lca.rootword.}{\small\tt play}\}\\
\hline
11 & AND: \{{\em v.text.}{\small\tt due}, {\em dep.path.u$<$pobj$<$prep$<$v}, {\em path.}{\small\tt to}\}\\
\hline
12 & AND: \{{\em v.POS\_gen.NOUN}, {\em dep.path.u$<$nsubj$<$LCA$>$attr$>$v}, {\em v.rootword.}{\small\tt cause}\}\\
\hline
13 & AND: \{{\em v.rootword.}{\small\tt cause}\}; OR: \{{\em dep.path.u$<$nsubjpass$<$LCA$>$xcomp$>$attr$>$v}, {\em dep.path.u$<$nsubjpass$<$LCA$>$prep$>$pobj$>$v}, {\em dep.path.u$<$dobj$<$LCA$>$prep$>$pobj$>$v}\}\\
\hline
14 & AND: \{{\em v.POS\_gen.VERB}, {\em v.POS.VBG}, {\em dep.path.u$<$nsubj$<$LCA$>$attr$>$acl$>$v}\}\\
\hline
15 & AND: \{{\em v.POS\_gen.VERB}, {\em dep.path.u$<$nsubj$<$v}, {\em u.POS\_gen.VERB}\}\\
\hline
16 & AND: \{{\em v.POS\_gen.VERB}, {\em edge.v.u.csubj}\}\\
17 & AND: \{{\em v.rootword.}{\small\tt die}, {\em dep.path.len.1.of$>$pobj$>$u}, {\em dep.path.u$<$pobj$<$prep$<$v}\}\\
\hline
18 & AND: \{{\em v.POS\_gen.VERB}, {\em dep.path.u$<$nsubj$<$LCA$>$xcomp$>$v}\}\\
\hline
19 & AND: \{{\em v.rootword.}{\small\tt result}, {\em v.child.}{\small\tt from}\}; OR: \{{\em dep.path.u$<$pcomp$<$prep$<$v}, {\em dep.path.u$<$pobj$<$prep$<$v}\}\\
\hline
20 & AND: \{{\em dep.path.u$<$nsubj$<$v}, {\em v.rootword.}{\small\tt result}, {\em v.child.}{\small\tt in}\}; OR: \{{\em u.POS\_gen.NOUN}, {\em u.POS\_gen.PROPN}\}; NEG: \{{\em v.child.}{\small\tt from}\}\\
\hline
21 & AND: \{{\em v.POS\_gen.NOUN}, {\em dep.path.u$<$pobj$<$prep$<$v}, {\em dep.path.len.1.for$>$pobj$>$u}\}; OR: \{{\em v.rootword.}{\small\tt consequence}, {\em v.rootword.}{\small\tt result}, {\em v.rootword.}{\small\tt effect}\}\\
\hline
22 & AND: \{{\em v.text.}{\small\tt because}, {\em dep.path.u$<$pobj$<$v}\}\\
\hline
23 & AND: \{{\em v.POS\_gen.VERB}, {\em v.POS.VBG}, {\em dep.path.u$<$nsubj$<$LCA$>$advcl$>$v}\}\\
\hline
24 & AND: \{{\em v.POS\_gen.NOUN}, {\em dep.path.u$<$nsubj$<$LCA$>$attr$>$v}, {\em v.rootword.}{\small\tt reason}\}\\
\hline
25 & AND: \{{\em v.POS\_gen.VERB}, {\em dep.path.u$<$nsubj$<$LCA$>$acomp$>$prep$>$pcomp$>$v}\}\\
\hline
26 & AND: \{{\em v.POS\_gen.VERB}, {\em dep.path.u$<$nsubj$<$LCA$>$acomp$>$xcomp$>$v}\}\\
\hline
27 & AND: \{{\em v.text.}{\small\tt because}, {\em dep.path.u$>$mark$>$v}\}\\
\hline
28 & AND: \{{\em v.text.}{\small\tt due}, {\em dep.path.u$<$pobj$<$pcomp$<$v}, {\em path.v.u.pcomp}, {\em path.v.u.pobj}, {\em ancestor.v.u}, {\em u.parent.dep.pobj}, {\em path.}{\small\tt to}\}\\
\hline
29 & AND: \{{\em v.POS\_gen.VERB}, {\em dep.path.u$<$nsubj$<$LCA$>$attr$>$relcl$>$v}, {\em u.POS\_gen.NOUN}\}\\
\hline
30 & AND: \{{\em v.POS\_gen.NOUN}, {\em u.POS\_gen.NOUN}, {\em dep.path.u$>$appos$>$v}\}; OR: \{{\em v.rootword.}{\small\tt inhibitor}, {\em v.rootword.}{\small\tt predictor}, {\em v.rootword.}{\small\tt marker}, {\em v.rootword.}{\small\tt cause}\}\\
\hline
31 & AND: \{{\em v.POS\_gen.NOUN}, {\em dep.path.u$<$nsubj$<$LCA$>$attr$>$v}\}; OR: \{{\em v.rootword.}{\small\tt inhibitor}, {\em v.rootword.}{\small\tt predictor}, {\em v.rootword.}{\small\tt marker}, {\em v.rootword.}{\small\tt cause}, {\em v.rootword.}{\small\tt complication}\}\\
\hline
32 & AND: \{{\em dep.path.u$<$nsubj$<$LCA$>$xcomp$>$attr$>$v}, {\em u.POS\_gen.NOUN}, {\em u.rootword.}{\small\tt cause}\}\\
\hline
33 & AND: \{{\em dep.path.u$<$csubjpass$<$LCA$>$xcomp$>$v}\}; NEG: \{{\em u.POS\_gen.PRON}\}\\
\hline
\end{tabular}
\caption{\label{tabCauseRules}The rules for identifying {\small\sf CAUSE} headwords ($v$ is a candidate causal trigger and $u$ is a candidate headword of a cause phrase)}
\end{table*}

\begin{table*}[h]\small
\begin{tabular}{cp{0.9\linewidth}}
\hline
\textbf{Rule} & \textbf{Rule}\\
\hline
1 & AND: \{{\em edge.v.u.dobj}\}\\
\hline
2 & AND: \{{\em v.POS\_gen.VERB}, {\em dep.path.u$>$amod$>$v}\} \\ 
\hline
3 & AND: \{{\em edge.v.u.nsubjpass}\} \\
\hline
4 & AND: \{{\em v.POS.VBN}, {\em edge.u.v.acl}, {\em u.POS\_gen.NOUN}, {\em v.POS\_gen.VERB}\} \\
\hline
5 & AND: \{{\em dep.path.u$>$prep$>$v}\}; OR:	\{{\em v.text.}{\small\tt because}, {\em v.text.}{\small\tt due}\}\\
\hline
6 & AND: \{{\em v.POS\_gen.VERB}, {\em v.rootword.}{\small\tt lead}, {\em dep.path.u$<$pobj$<$prep$<$v}, {\em path.}{\small\tt to}\} \\
\hline
7 & AND: \{{\em v.POS\_gen.NOUN}, {\em dep.path.u$<$pobj$<$prep$<$v}, {\em path.of}\};\newline OR: \{{\em v.rootword.}{\small\tt cause}, {\em v.rootword.}{\small\tt reason}, {\em v.child.prep.}{\small\tt by}, \newline{\em v.child.agent.}{\small\tt by}\} \\
\hline
8 & AND: \{{\em v.POS\_gen.VERB}, {\em v.rootword.}{\small\tt contribute}, \newline{\em dep.path.u$<$pobj$<$prep$<$v}, {\em path.}{\small\tt to}\} \\
\hline
9 & AND: \{{\em v.rootword.}{\small\tt role}, {\em dep.path.u$<$pobj$<$prep$<$v}, {\em path.}{\small\tt in}\} \\
\hline
10 & AND: \{{\em dep.path.u$>$advcl$>$mark$>$v}, {\em v.text.}{\small\tt because}\} \\
\hline
11 & AND: \{{\em v.rootword.}{\small\tt result}, {\em v.child.}{\small\tt from}, {\em dep.path.u$<$nsubj$<$v}\}\\
\hline
12 & AND: \{{\em v.rootword.}{\small\tt role}, {\em dep.path.u$<$pcomp$<$prep$<$v}, {\em path.}{\small\tt in}\}\\
\hline
13 & AND: \{{\em v.rootword.}{\small\tt result}, {\em v.child.}{\small\tt in}\}; OR: \{{\em dep.path.u$<$pcomp$<$prep$<$v}, {\em dep.path.u$<$pobj$<$prep$<$v}\}\\
\hline
14 & AND: \{{\em v.POS\_gen.NOUN}, {\em dep.path.u$<$pobj$<$prep$<$v}, {\em dep.path.len.1.for$>$pobj$>$u}, {\em v.rootword.}{\small\tt cause}\}\\
\hline
15 & AND: \{{\em v.POS\_gen.NOUN}, {\em dep.path.u$<$pcomp$<$prep$<$v}, {\em dep.path.len.1.of$>$pcomp$>$u}\}; OR: \{{\em v.rootword.}{\small\tt cause}, {\em v.rootword.}{\small\tt reason}\}\\
\hline
16 & AND: \{{\em v.POS.VBN}, {\em dep.path.u$<$nsubj$<$LCA$>$attr$>$acl$>$v}, {\em v.POS\_gen.VERB}\}\\
\hline
17 & AND: \{{\em edge.u.v}, {\em lca.rootword.}{\small\tt be}, {\em v.text.}{\small\tt due}, {\em u.copula\_verb\_with\_object}\}\\
\hline
18 & AND: \{{\em v.rootword.}{\small\tt die}, {\em dep.path.u$<$nsubj$<$v}, {\em v.child.}{\small\tt of}\}\\
\hline
19 & AND: \{{\em v.POS\_gen.NOUN}, {\em dep.path.u$<$pobj$<$prep$<$v}, {\em dep.path.len.1.for$>$pobj$>$u}, {\em v.rootword.}{\small\tt reason}\}\\
\hline
20 & AND: \{{\em v.text.}{\small\tt due}, {\em dep.path.u$<$nsubj$<$LCA$>$acomp$>$v}, {\em lca.rootword.}{\small\tt be}\}\\
\hline
21 & AND: \{{\em v.text.}{\small\tt due}, {\em dep.path.u$>$amod$>$v}\}\\
\hline
\end{tabular}
\caption{\label{tabEffectRules}The rules for identifying {\small\sf EFFECT} headwords ($v$ is a candidate causal trigger and $u$ is a candidate headword of an effect phrase)}
\end{table*}

\end{document}